\documentclass[11pt,a4paper]{article}

\usepackage[hyperref]{emnlp2020}
\usepackage{times}
\usepackage{latexsym}

\usepackage{microtype}

\aclfinalcopy 
\def\aclpaperid{***} 

\newcommand\blfootnote[1]{%
  \begingroup
  \renewcommand\thefootnote{}\footnote{#1}%
  \addtocounter{footnote}{-1}%
  \endgroup
}

\newcommand{\thiscorpus}{\textsc{GGPOnc}}
\usepackage{pifont}
\newcommand{\cmark}{\ding{51}}%
\newcommand{\xmark}{\ding{55}}%

\usepackage{listings}
\usepackage[utf8]{inputenc}
\usepackage{paralist}
\usepackage{multirow}
\usepackage{graphicx}

\title{\thiscorpus: A Corpus of German Medical Text with Rich Metadata Based on Clinical Practice Guidelines}
\author{
    Florian Borchert$^{1,4*}$,
    Christina Lohr$^{2,5*}$,
    Luise Modersohn$^{2,5*}$,
    Thomas Langer$^3$,\\
    \textbf{Markus Follmann$^3$,
    Jan Philipp Sachs$^1$,
    Udo Hahn$^{2,5}$, and
    Matthieu-P. Schapranow$^{1,4}$}\\
    $^1$Digital Health Center, Hasso Plattner Institute, University of Potsdam, Germany \\
    \texttt{\{firstname.lastname\}@hpi.de}\\
    $^2$Jena University Language \& Information Engineering (\textsc{Julie}) Lab\\Friedrich Schiller University Jena, Germany\\
    \texttt{\{firstname.lastname\}@uni-jena.de}\\
    $^3$German Guideline Program in Oncology, German Cancer Society, Berlin, Germany\\
    \texttt{\{lastname\}@krebsgesellschaft.de}\\
    $^4$\textsc{HiGHmed} Consortium of the German Medical Informatics Initiative\\
    $^5$\textsc{SMITH} Consortium of the German Medical Informatics Initiative\\
}

\date{}

\begin{document}
\maketitle

\begin{abstract}
The lack of publicly accessible text corpora is a major obstacle for progress in natural language processing. For medical applications, unfortunately, all language communities other than English are low-resourced.
In this work, we present \thiscorpus\ (German Guideline Program in Oncology NLP Corpus), a freely distributable German language corpus based on clinical practice guidelines for oncology.
This corpus is one of the largest ever built from  German medical documents. Unlike clinical documents, clinical guidelines do not contain any patient-related information and can therefore be used without data protection restrictions. Moreover, \thiscorpus\ is the first corpus for the German language covering diverse conditions in a large medical subfield 
and provides a variety of metadata, such as literature references and evi\-dence levels.
By applying and evaluating existing medical information extraction pipelines for German text, we are able to draw comparisons for the use of medical language to other corpora, medical and non-medical ones.
\end{abstract}
\section{Introduction}
\label{sec:intro}
\blfootnote{$^*$Authors marked by $^*$ equally share first authorship.}The synthesis of validated experience in the form of Clinical Practice Guidelines (CPGs) serves as a basis for evidence-based decision making in clinical practice. To leverage the knowledge in CPGs for clinical decision support systems, e.g., for integration with electronic health records or automated evaluation of adherence to these guidelines, machine-readable versions of CPGs are necessary. However, CPGs today are disseminated mostly as free-text documents, with few formal elements. 
Thus, Natural Language Processing (NLP) might be helpful to automatically extract information from these unstructured texts and transform them into a structured, or even machine-executable, format. As CPGs are also specific to their country of origin, they are usually formulated in the respective native language, so NLP technology has to be adapted properly.

A major reason for the progress in NLP research in the past years is the public
availability of large text corpora (and attached metadata). 
Yet, for documents originating from a clinical context, the protection of personal information is a major requirement imposed by legal privacy regulations. Some research initiatives, e.g., \textsc{i2b2} \cite{Uzuner11}, \textsc{Mimic-III} \cite{Johnson16},
the Shared Task of Social Media for Health \cite{weissenbacher-etal-2019-overview}, or \textsc{Clef eHealth} \cite{Lorraine20CLEFeHealth} make de-identified
clinical text document collections available under the conditions of 
Data Use Agreements (DUA). Besides, databases of biomedical research articles like \textsc{PubMed} provide an abundant amount of examples for medical language. 
However, with only few exceptions, such easily accesssible text corpora are hardly available for the German \cite{LOHR18.701} and other non-English languages.
Today, there is no viable solution for sharing even de-identified clinical texts in Germany. 
In effect, not only are large-scaled
research datasets missing but also pretrained language models for German medical language, such as an equivalent of BioBERT \citep{lee2020biobert}.

To address \textit{(1)} the lack of available German medical text resources for NLP research, and \textit{(2)} the need for machine-readable CPGs, we constructed a corpus based on a set of German CPGs for oncology.
The \textit{German Guideline Program in Oncology (GGPO)} \cite{llonko}, operated by the Association of the Scientific Medical Societies in Germany, the German Cancer Society and the German Cancer Aid, is in a unique position to enable this research, as their guidelines are also provided via a mobile app \cite{Seufferlein2019}. Hence, this data set is already available in a semi-structured format with rich, formatted metadata, resulting in a much higher data quality than data extracted a posteriori from \textsc{PDF} versions of the guidelines. 

An excerpt from the XML version of \thiscorpus\ is depicted in Listing \ref{lst:xml_sample}. Other than clinical text relating to individual patients, we deal with scientific medical text that does not contain any privacy-sensitive data requiring de-identification. This way, we can provide access to \thiscorpus\ for other researchers via a DUA.\footnote{For instructions on how to get access to the data, see:  
{\url{https://www.leitlinienprogramm-onkologie.de/projekte/ggponc-english/}}}

\definecolor{darkblue}{rgb}{0,0,0.5} 
\definecolor{darkred}{rgb}{0.5,0,0} 

\lstset{
    captionpos=b,
    language=xml,
    tabsize=1,
    caption={Text snippet from the XML version of the underlying GGPO corpus. Documents are structured into sections which can contain multiple recommendations. CPG recommendations can carry a multitude of metadata elements, as well as a concise text statement. Additional background text segments may contain more detailed information.},
    label=lst:xml_sample,
    frame=single,
    rulesepcolor=\color{gray},
    xleftmargin=0pt,
    framexleftmargin=0pt,
    keywordstyle=\color{blue}\bf,
    commentstyle=\color{gray},
    stringstyle=\color{darkred},
    numberstyle=\tiny,
    breaklines=true,
    showstringspaces=false,
    basicstyle=\footnotesize\ttfamily,
    literate=%
  {Ö}{{\"O}}1
  {Ä}{{\"A}}1
  {Ü}{{\"U}}1
  {ß}{{\ss}}1
  {ü}{{\"u}}1
  {ä}{{\"a}}1
  {ö}{{\"o}}1,
  emph={corpus,document,section, text, name, recommendation, number, recommendation_grade, edit_state, type_of_recommendation, level_of_evidences, loe, literature_references, litref,strength_of_consensus, vote, recommendation_creation_date},
  emphstyle={\color{darkblue}}
}
\vspace{15pt}
\begin{minipage}{0.95\linewidth}
\begin{lstlisting}
<document>
 <section>
  <name>Risikofaktoren</name>
  <section>
   <name>Helicobacter pylori</name>
   <recommendation>
    <recommendation_creation_date value="2019-01-01"/>
    <recommendation_grade value="B"/>
    <!-- more metadata -->
    <text>Die H. pylori-Eradikation mit dem Ziel der Magenkarzinom-prävention sollte bei den folgenden Risikopersonen durchgeführt werden (siehe Tabelle unten).</text>
   </recommendation>
   <text>Das Magenkarzinom ist eine multifaktorielle Erkrankung, bei der die Infektion mit H. pylori den wichtigsten Risikofaktor darstellt. Seit 1994 ist H. pylori durch die Weltgesund-heitsorganisation als Klasse I Karzinogen anerkannt und wurde 2009 als solches bestätigt <litref id="65327"/>.</text>
  </section> <!-- more sections -->
 </section>
</document> <!-- more documents -->

\end{lstlisting}
\end{minipage}
\section{Related work}
\label{sec:related}


Due to legal data protection measures, German-language clinical text corpora are extremely rare and existing ones almost impossible to (re)use. Typi\-cally, accessibility is restricted to research staff only within the lifetime of a project and blocked for the outside world. 
In Table~\ref{tab:existing_corpora}, we list, to the best of our knowledge, all existing German-language clinical research text corpora which have been described in scientific publications up until now. With only few exceptions, these corpora are small and mostly limited to a specific medical discipline or clinical division.
In addition to these pure clinical documents, other document types are also interesting for the NLP community, e.g., CPGs, which are available for a wide range of conditions.

\begin{table*}[!ht]
\centering
\caption{Overview of existing text corpora of German clinical language. For \thiscorpus, we report the number of guidelines with the number of their individual text segments in brackets.}
\vspace{-2mm}
\begin{tabular} {| p{7.8cm} | r | r | r | c |}
\hline
\textbf{Corpus / Data} &
\textbf{Documents} &
\textbf{Sentences} &
\textbf{Tokens} &
\textbf{Available}\\
\hline \hline
\textsc{FraMed}: clinical reports and medical textbook snippets  \cite{Wermter04}
&
-- &
6k &
100k &
\xmark\\
\hline
Reports from five medical domains \newline \cite{Fette12} &
544 &
-- &
-- &
\xmark\\
\hline
Radiology reports \cite{Bretschneider13identifying}
&
174 &
4k &
28k &
\xmark\\
\hline
Transthoracic echocardiography reports  \newline \cite{Toepfer15} &
140 &
-- &
-- &
\xmark\\
\hline
Operative reports (surgery) \cite{Lohr16} &
450 &
22k &
266k &
\xmark\\
\hline
Discharge summaries from a dermatology department \cite{Kreuzthaler16}
&
1,696 &
-- &
-- &
\xmark\\
\hline
Discharge summaries and clinical notes from nephrology domain \cite{Roller16} &
1,725 &
28k &
158k &
\xmark\\
\hline
Discharge summaries and clinical notes from nephrology domain \cite{Cotik16} &
183 &
2k &
13k &
\xmark\\
\hline
X-ray reports \cite{Krebs17} &
3,000 &
-- &
-- &
\xmark\\
\hline
\textsc{3000PA}: internistic and ICU discharge summaries
&
$\approx$~3,000 &
-- &
-- &
\xmark\\
\textsc{3000PA Jena Part} \cite{Hahn18}
&
1,006 &
170k &
1,421k &
\xmark\\\hline
\textsc{JSynCC}: case examples from medical textbooks \cite{LOHR18.701} (v1.1) &
903 &
29k &
368k &
{\cmark}\\\hline
Mixed-domain, -section, and -document type  &
60  (400–600  &
--&
$\approx$~6k &
{\xmark}\\
\textsc{Assess CT} corpus
\cite{MinarroGimenez19} &
chars each) &
&
&
\\
\hline
Discharge summaries with osteoporosis diagnosis \cite{konig2019knowledge} &
1,982&
--&
2,001k&
\xmark\\
\hline
Technical-Laymen Corpus: social media samples (Stomach-Intestines, Kidney) \cite{seiffe-etal-2020-witchs} &
4,000 &
--&
438k &
{\cmark}\\
\hline
\textbf{\thiscorpus\ -- recommendations } &
25 (4,348)&
7k &
132k &
{\cmark}\\
\hline
\textbf{\thiscorpus\ -- complete corpus} &
25 (8,418)&
60k &
1,340k &
{\cmark}\\
\hline
\end{tabular}
\label{tab:existing_corpora}
\end{table*}

CPGs as a target for automated text analytics have been much less utilized compared to other scientific publications and clinical documents. Most of that work 
took place in the context of for\-ma\-li\-zing CPGs as computer-interpretable guidelines \cite{peleg2013computer}.
\citet{Bouffier2007} describe an approach to fill in a semi-structured \textit{Guideline Elements Model} template by segmenting unstructured guidelines using linguistic patterns. An evaluation was run on 18~French guidelines.
\citet{Serban2007} describe the extraction and instantiation of linguistic templates for guideline formalization, evaluated on a Dutch guideline for breast cancer treatment. 
German CPGs were the focus of \citet{Becker2017} who adapted \textsc{Apache cTakes} to detect German UMLS concepts and evaluated their approach on a single German breast cancer guideline.
\citet{Zadrozny17} outline a system which identifies contradictions and disagreements in English CPGs.


Some authors have focused on extracting more task-specific information, such as activities \cite{kaiser2010identifying}, process structures \cite{Wenzina2013, Zhu2013, HematialamZ17} or negation triggers \cite{Gindl2008}. \citet{taboada2013} apply a pipeline of open-source tools for parsing CPGs, Named Entity Recognition (NER) tagging and relation extraction in a case study with 171 sentences from CPGs. Most of the aforementioned approaches work with relatively small annotated corpora and English language, only. Recently, \citet{fazlic2019} use LSTMs and fuzzy rules to extract \textit{``action takers"}, \textit{``symptoms"}, \textit{``actions"} and \textit{``purposes"} from CPGs, recognize recommendations and predict the grade of recommendation. The authors use a data set extracted from PDF versions of 45~guidelines with 1,020~recommendations.

\begin{table*}[!ht]
  \vspace*{-4pt}
 \caption{Metadata elements of recommendations of \thiscorpus\ }
  \centering
  \begin{tabular}{| l | p{10cm} |}
   \hline
    \textbf{Attribute}     & \textbf{Description }\\ \hline \hline
    Recommendation creation date & Date the recommendation was first introduced\\ \hline
    Type of recommendation & Evidence-based or consensus-based statement or recommendation\\ \hline
    Recommendation grade & A (strong recommendation)  \\ 
    & B (recommendation) \\ 
    & 0 (weak recommendation / option)  \\  \hline
    Strength of consensus & Strong Consensus  \\ 
    & Consensus \\ 
    & Approved by majority \\ 
    & No consensus  \\ \hline
    Total vote in percentage & Percentage of approval among the expert committee \\  \hline
    Literature references & List of evidence backing up the recommendation \\  \hline
    Expert opinion & Yes or absent \\  \hline
    Level of evidence & According to Oxford,
    SIGN,
    or GRADE
    \\  \hline
    Edit state & State (checked, new, modified) \&  note regarding guideline updates \\  \hline
  \end{tabular}
  \vspace*{-4pt}
  \label{tab:metadata}
\end{table*}

Some larger corpora of CPGs for the English language exist already. \citet{Hussain2009} present the Yale Guideline Re\-commendation Corpus (YGRC), a sample of 1,275~guideline recommendations extracted from the \textit{National Guideline Clearinghouse} (NGC). Their work revealed inconsistencies in writing style and reporting of the strength of recommendations. 
Using a subset of YGRC, \citet{GadEl-Rab2017} present a rule-based approach to detect procedures and drug recommendations. \citet{READ16.521} describe the CREST corpus, consisting of 4,029~recommendations from 170~guidelines annotated with their respective recommendation strength and report a total number of 8,138~types within the recommendations.
%
Large corpora of CPGs lend themselves to mining the state-of-the-art knowledge in a medical subfield. For instance, \citet{Leung2015a} identify comorbidities by analyzing pairs of co-occurring conditions, using a corpus of 268~NGC guideline summaries.
\citet{Leung2016} find drug-disease relations via named entity recognition using a corpus of 377~NGC guideline summaries. The  relations are compared to structured drug product labels to assess their overlap.

In summary, our work is most similar to the CREST corpus \citep{READ16.521}, in the sense that we provide a corpus based on CPGs consisting of medical text and metadata. However, while the number of recommendations in \thiscorpus\ is comparable to CREST, the amount of structured metadata and background text in our corpus is much larger (see Section \ref{sec:characteristics}). Also, our corpus contains German text, addressing a scenario where available resources are much scarcer (see Table~\ref{tab:existing_corpora}). While \citet{Becker2017} also apply NLP to German CPGs, we consider a large superset of the CPGs used in their work and provide access to our data as a preprocessed and analyzed text corpus. 






\bgroup
\begin{table*}[h!]
 \caption{
 Details of \thiscorpus. We report the number of text \textbf{Seg}ments (plain text files), \textbf{Rec}ommendations and literature \textbf{Ref}erences. The numbers of \textbf{Sent}ences, \textbf{Tokens} and \textbf{Types} refer to the pure textual content of the corpus, excluding any meta-data and headings. Annotated parts of the corpus are marked with * (see also Section \ref{sec:evalanno}).
 }

  \centering
  \setlength{\tabcolsep}{1.7mm}
  \begin{tabular}{| r | l | r | r | r | r | r | r |}
  \hline
    &  \textbf{Guideline} & \textbf{Seg.} & \textbf{Rec.} & \textbf{Sent.} & \textbf{Tokens} & \textbf{Types} & \textbf{Ref.}\\ \hline \hline

1 & Palliative medicine* & 696 & 445 & 5,956 & 134,489 & 15,795 & 3,065 \\ \hline 
2 & Lung cancer* & 666 & 313 & 4,251 & 93,324 & 12,756 & 2,344 \\ \hline 
3 & Breast cancer & 685 & 362 & 4,127 & 93,128 & 12,660 & 2,824 \\ \hline 
4 & Supportive therapy & 823 & 337 & 4,224 & 90,711 & 12,411 & 2,401 \\ \hline 
5 & Bladder cancer & 355 & 225 & 3,872 & 85,299 & 11,347 & 2,521 \\ \hline 
6 & Colorectal cancer* & 569 & 290 & 3,176 & 71,416 & 9,644 & 2,580 \\ \hline 
7 & Prostate cancer & 307 & 221 & 3,090 & 67,900 & 9,418 & 2,119 \\ \hline 
8 & Malignant melanoma & 297 & 167 & 2,715 & 60,354 & 9,318 & 1,256 \\ \hline 
9 & Prevention of skin cancer & 288 & 119 & 2,354 & 55,965 & 9,140 & 952 \\ \hline 
10 & Actinic keratosis and SCC of the skin* & 199 & 74 & 2,590 & 54,073 & 6,861 & 1,278 \\ \hline 
11 & Stomach cancer & 246 & 142 & 2,328 & 50,836 & 8,156 & 1,670 \\ \hline 
12 & Endometrial cancer & 317 & 173 & 1,999 & 50,056 & 8,154 & 1,340 \\ \hline 
13 & Cervical cancer* & 341 & 115 & 2,168 & 49,422 & 8,164 & 1,127 \\ \hline 
14 & Prevention of cervix cancer* & 302 & 103 & 2,055 & 48,676 & 7,989 & 1,391 \\ \hline 
15 & Renal cell cancer* & 276 & 122 & 2,118 & 48,013 & 8,202 & 1,496 \\ \hline 
16 & Testicular tumors & 315 & 163 & 1,917 & 43,726 & 6,774 & 1,412 \\ \hline 
17 & Oesophageal cancer* & 172 & 91 & 1,611 & 35,710 & 6,680 & 1,026 \\ \hline 
18 & Laryngeal cancer & 189 & 118 & 1,525 & 35,519 & 6,841 & 681 \\ \hline 
19 & Chronic lymphocytic leukemia (CLL)* & 290 & 138 & 1,410 & 34,470 & 5,682 & 725 \\ \hline 
20 & Hodgkin lymphoma* & 253 & 167 & 1,489 & 31,876 & 5,245 & 889 \\ \hline 
21 & Hepatocellular cancer (HCC)* & 157 & 88 & 1,296 & 27,852 & 5,704 & 803 \\ \hline 
22 & Malignant ovarian tumors & 193 & 94 & 1,136 & 25,807 & 5,110 & 1,013 \\ \hline 
23 & Psycho-oncology* & 121 & 47 & 778 & 19,270 & 4,127 & 835 \\ \hline 
24 & Pancreatic cancer & 294 & 158 & 857 & 16,871 & 3,670 & 1,154 \\ \hline 
25 & Oral cavity cancer* & 111 & 76 & 630 & 15,438 & 3,376 & 1,026 \\ \hline 
\hline & \textbf{Annotated Part} & \textbf{4,153} & \textbf{2,069} & \textbf{29,528} & \textbf{664,029} & \textbf{50,732} & \textbf{18,585} \\ \hline 
& \textbf{Full Corpus} & \textbf{8,414} & \textbf{4,348} & \textbf{59,672} & \textbf{1,340,201} & \textbf{76,252} & \textbf{37,928} \\ \hline


  \end{tabular}
  \label{tab:corpus_stats}

\end{table*}
\egroup

\begin{table*}[]
 \caption{Comparison of \thiscorpus\ with 3000PA (Jena part), \textsc{JSynCC}, German \textsc{PubMed} abstracts of case reports and two non-clinical corpora (German Wikipedia articles of wars (\textsc{WikiWarsDE}) and news articles from the \textsc{Krauts} corpus)}
  \centering
\begin{tabular}{|l r|r|r|r|r|r|r|r|}
\hline
& & \multicolumn{2}{|c|}{\textbf{\thiscorpus}} &
\multicolumn{3}{|c|}{\textbf{Clinical}} & \multicolumn{2}{|c|}{\textbf{Non-Clinical}}  \\ \cline{3-9}
& & \textbf{Complete} & \textbf{Recom.} & \textbf{3000PA-J} & \textbf{\textsc{JSynCC}} & \textbf{\textsc{PubMed}} & \textbf{\textsc{Wiki}} & \textbf{\textsc{Krauts}} \\ \hline \hline
\multicolumn{2}{|l|}{Documents} & 8,418 & 4,348 & 1106 & 903 & 336 & 22 & 142 \\ \hline
\multicolumn{2}{|l|}{Sentences} & 60k & 7k & 171k & 29k & 3k & 5k & 1k \\ \hline
\multicolumn{2}{|l|}{Tokens} & 1,340k & 132k & 1,421k & 368k & 43k & 96k & 31k \\ \hline
\multicolumn{2}{|l|}{Tokens / Sentence} & 22.5 & 19.0 & 8.8 & 12.5 & 16.5 & 20.9 & 25.3 \\ \hline
UMLS* & (\%) & 6.42 & 8.93 & 8.72 & 5.71 & 7.59 & 0.75 & 0.02 \\ \hline
\hspace{10pt}ANAT & (\%)& 0.45 & 0.48 & 1.78 & 1.11 & 0.79 & 0.04 & 0.09 \\ \hline
\hspace{10pt}CHEM & (\%)& 0.82 & 1.01 & 1.08 & 0.41 & 0.59 & 0.04 & 0.07 \\ \hline
\hspace{10pt}DEVI & (\%)& 0.12 & 0.17 & 0.20 & 0.55 & 0.18 & 0.06 & 0.04 \\ \hline
\hspace{10pt}DISO & (\%)& 1.42 & 2.02 & 2.96 & 1.21 & 2.80 & 0.08 & 0.13 \\ \hline
\hspace{10pt}LIVB & (\%)& 1.07 & 1.32 & 0.38 & 0.35 & 0.82 & 0.38 & 0.37 \\ \hline
\hspace{10pt}PHYS & (\%)& 0.37 & 0.43 & 0.76 & 0.60 & 0.50 & 0.12 & 0.10 \\ \hline
\hspace{10pt}PROC & (\%)& 2.18 & 3.50 & 1.56 & 1.49 & 1.90 & 0.01 & 0.12 \\ \hline
Genes & (\%)& 1.28 & 1.41 & 2.21 & 0.87 & 0.97 & 0.94 & 0.55 \\ \hline
TNM & (\%) & 0.19 & 0.37 & 0.07 & 0.07 & 0.04 & 0.003 & 0 \\ \hline
Stop words & (\%)& 34.05 & 35.53 & 20.37 & 32.96 & 34.51 & 34.65 & 24.24 \\ \hline
\end{tabular}
  \label{tab:tagging_results}
\end{table*}

\section{Methods}
\label{sec:contribution}

\subsection{Data Collection}

In order to assemble the corpus of German CPGs, we acquired semi-structured \textsc{JSON} versions of the guidelines from the REST API of the Content Ma\-nage\-ment System (CMS) that serves the backend for the mobile app provided by the GGPO. The data was subsequently transformed from \textsc{JSON} to an \textsc{XML} format. We preserved the document structure (chapters and sections), as well as recommendation metadata and literature references. An example of the resulting \textsc{XML} format can be found in Listing~\ref{lst:xml_sample}. The metadata elements are described in Table \ref{tab:metadata}. Literature references are included with an ID number which can be resolved to a citation in the provided literature index file.

The guidelines distinguish between re\-com\-men\-da\-tions and background texts; we preserved this distinction in the corpus. In general, the recommendations tend to be concise statements related to a particular clinical question. For evidence-based re\-com\-men\-da\-tions, literature references and evidence levels are included. The background texts provide the reasoning behind the recommendations and a summary of the evidence underlying the recommendations, again backed by literature references. 

\subsection{Automated Annotation}

Besides the \textsc{XML} version of the corpus, we crea\-ted plain text versions of all recommendations of background text parts to facilitate processing by existing NLP pipelines. For preprocessing, like sentence splitting and tokenization, we used 
\textsc{JCoRe} \cite{Hahn16} (i.e., \textsc{Uima}-based) pipelines and \textsc{FraMed} \cite{Wermter04} models which were developed for German clinical text.

We also employed the \textsc{JuFit} tool (v1.1) \cite{Hellrich15}, a filter for UMLS to create a dictionary of all German words from the UMLS \cite{Bodenreider04} (version 2019AB)\footnote{\mbox{\url{https://www.nlm.nih.gov/research/umls/}}} and the Semantic Groups ANAT (Anatomical Structure), CHEM (Chemicals \& Drugs), DEVI (Devices), DISO (Disorders), LIVB (Living Beings), PHYS (Physiology), and PROC (Procedures) (without advanced \textsc{JuFit} rules).
We have chosen only these six out of the full set of 15 UMLS Semantic Groups because we used similar categories in the named entity recognition tasks and also wanted to avoid cognitive overloading of the human annotators.



Finally, we screened TNM expressions\footnote{The UICC TNM system is a classification scheme for malignant tumors, see \url{https://www.uicc.org/resources/tnm}} which were extracted using a rule-based approach implemented with the \textsc{Python} library \textsc{spaCy}. This part was originally developed for German pathology reports in the context of the \textsc{HiGHmed} consortium of the Medical Informatics Initiative of Germany.
TNM expressions and genes were specifically chosen for their relevance in cancer treatment. 


\section{Results}\label{sec:results}
\vspace{-2pt}
\subsection{Corpus Characteristics}
\label{sec:characteristics}

\enlargethispage{1em}
In total, 25 GPGs with 8,414 text segments were extracted from the CMS comprising the first version of \thiscorpus. In Table \ref{tab:corpus_stats}, we give an overview of the CPGs in terms of the number of tokens and types, as well as the number of literature references. We also report the total number of recommendations and background text segments, since they serve as the units of analysis for our automated annotation pipelines. The CPGs cover a wide range of indications and anatomical locations. They also differ significantly in their extent, e.g., there is much more text for broad topics, such as palliative medicine, or indications with many treatment options, such as lung cancer.
Of the approximately~38k literature references in the corpus, around~20k are unique with roughly~9k explicit links to \textsc{PubMed}. We provide bibliographic details on these references alongside the corpus to facilitate research on the relationships between CPGs and the underlying medical evidence.

Table \ref{tab:tagging_results} contains the automated named entity extraction results. Their quality and interpretation in comparison to other German (clinical and non-clinical) text corpora will be discussed in the next section. 
The whole corpus consists of:
 \vspace*{1pt}
 
\begin{itemize}
    \item a single \textsc{XML} file, including the document structure and all mentioned metadata,
    \vspace*{-5pt}
    \item a file for the complete literature index,
     \vspace*{-5pt}
    \item individual plain text versions of the text segments, sentences, and tokens,
     \vspace*{-5pt}
    \item automatically created entity annotations and a subset of manually corrected annotations in standoff format.\vspace{5pt}
\end{itemize}
\enlargethispage{1em}

As CPGs are subject to a regular update cycle, we are able to automatically redo the data acquisition process in the future in order to provide a historical view on the guideline development.
%






\subsection{Comparison with Other German Medical and Non-Medical Corpora}

We analyze the characteristics of \thiscorpus\ by comparing the entity matches with three German medical text corpora, namely version 1.1 of the \textsc{JSynCC} corpus (case examples from clinical text books) \cite{LOHR18.701}, the Jena Part of the 3000PA corpus (1106 German discharge summaries) \cite{Hahn18},\footnote{Based on the approval by the local ethics committee (4639-12/15) and the data protection officer of Jena University Hospital discharge summaries were extracted from the HIS of the Jena University Hospital and further transformed.}
and abstracts of German case reports from \textsc{PubMed}. In addition, we compare the results to out-of-domain corpora, namely German \textsc{Wikipedia} articles of wars (\textsc{WikiWarsDE}) \cite{wikiwarsde} and news articles from the \textsc{Krauts} corpus \cite{strotgen-etal-2018-krauts}. The results are depicted in Table \ref{tab:tagging_results}.

The fraction of stop words is comparable across all medical text corpora, as is the fraction of tokens that map to UMLS concepts. As expected, the guideline recommendations contain more medical terms per token than the background text. Compared to the clinical corpora, the CPGs have more instances of the class \textit{Living Beings}, as they often describe treatment recommendations for certain populations. Notably, the average sentence length is much greater in the clinical guidelines, and in particular in the background text, pointing at the more scientific style of writing prevalent in the guidelines as compared to clinical narratives. TNM expressions occur much more frequently in \thiscorpus, which can be attributed to its focus on the oncology domain. 

\begin{figure}[]
  \centering
  \includegraphics[height=125px, trim={0 0 0 120px},clip]{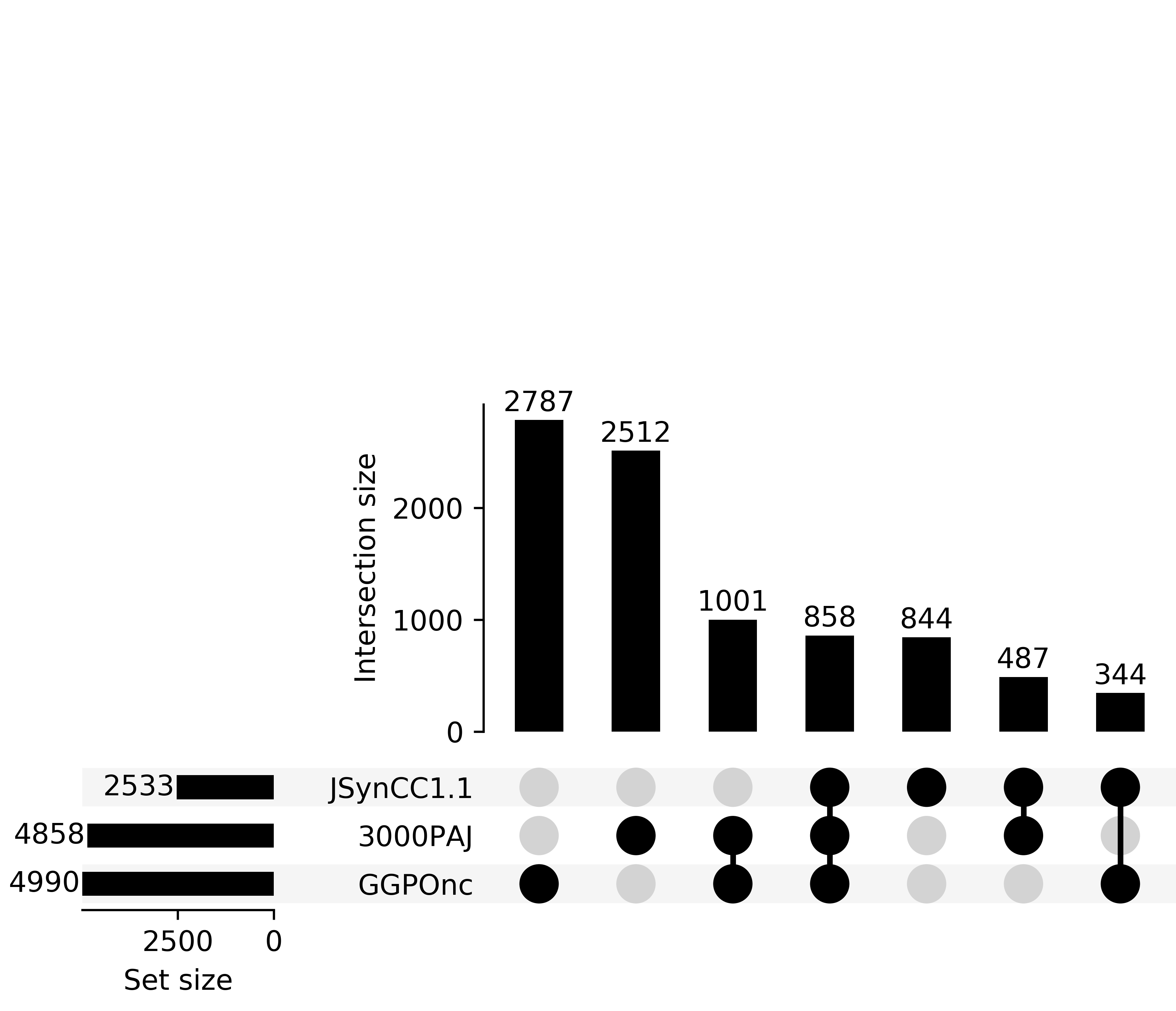}
  \caption{\textsc{UpSet} \cite{upset} visualization of the intersection of distinct UMLS concepts in \textsc{JSynCC} 1.1, 3000PA (Jena part) and \textsc{GGPOnc}. Each vertical bar indicates the size of one intersecting subset of UMLS terminology shared between the corpora, whereas the horizontal bars denote the total number of distinct UMLS concepts per corpus.}
  \label{fig:upset_intersection}
\end{figure}

Both out-of-domain corpora contain only small amounts of  
UMLS concepts (apart from the semantic class \textit{Living Beings}), which indicates a high precision of our entity tagging approach.
In Figure~\ref{fig:upset_intersection} we visualize the overlap of unique medical concepts from UMLS found in each of the corpora.

While there is a significant overlap between \thiscorpus\ and the clinical corpora, a major fraction of concepts is unique to each corpus.
These results suggest that our corpus combined with other clinical text corpora can provide a more comprehensive view on the use of medical language, in general, than each of the corpora alone.

\begin{table*}[h!]
\centering
\caption{Pair-wise average F1-score and standard deviation ($\sigma$) for instance and token based inter-annotator-agreement (IAA), precision and recall per entity class. Genes had to be excluded from IAA analysis due to the large number of false positives.}
\begin{tabular}{|ll|r|r|r|r|r|r|}
\hline

& & \multicolumn{2}{|c|}{\textbf{IAA Instances}} & \multicolumn{2}{|c|}{\textbf{IAA Token}} & \multirow{ 2}{*}{\textbf{Precision}}  & \multirow{ 2}{*}{\textbf{Recall}} \\ \cline{3-6}
& & \textbf{avg. F-score} & \textbf{$\sigma$} & \textbf{avg. F-score} & \textbf{$\sigma$} & & \\ \hline \hline
UMLS Anatomy & (ANAT)         & .718 & .122 & .720 & .125 & .872 & .571 \\ \hline

UMLS Chemicals & (CHEM)       & .839 & .045 & .850 & .041 & .917 & .600 \\ \hline

UMLS Devices & (DEVI)         & .414 & .366 & .409 & .368 & .465 & .209 \\ \hline

UMLS Disorders & (DISO)       & .747 & .097 & .773 & .096 & .919 & .453 \\ \hline

UMLS Living Being & (LIVB)    & .848 & .066 & .846 & .067 & .985 & .698 \\ \hline

UMLS Physiology & (PHYS)      & .534 & .183 & .576 & .177 & .607 & .310 \\ \hline

UMLS Procedures & (PROC)      & .706 & .099 & .727 & .100 & .944 & .506 \\ \hline
\multicolumn{2}{|l|}{TNM (rule based)}            & .820 &  .073 & .749 & .120 & .965 & .881 \\ \hline \hline
\multicolumn{2}{|l|}{Overall w/o Genes}           & \textbf{.742} & \textbf{.094} & \textbf{.758} & \textbf{.094} & \textbf{.945} & \textbf{.528} \\ \hline \hline
\multicolumn{2}{|l|}{Genes}                       & - & - & - &  - & .022 & .589 \\ \hline
\end{tabular}
\label{tab:evaluation}
\end{table*}


\subsection{Evaluation of Annotation Results}\label{sec:evalanno}

The automatic annotations for a subset of the CPGs have been independently reviewed by human experts (4~students of medicine, all of them passed their first medical exam, supervised by a medical doctor) using the \textsc{Brat} annotation tool \cite{Stenetorp12}.
Due to restricted resources for manual annotation work, we decided to evaluate on a subset of 13 (full) guidelines (see Table \ref{tab:corpus_stats}), which amounts to half of the corpus. The CPGs were chosen such that they cover a diverse range of topics and percentages of token matches, with a rather high rate of around~6--8\% results per token for HCC as opposed to a lower rate of roughly~5--6\% for CLL and psycho-oncology. 
Additional guidelines were chosen for manual annotation based on project requirements.

We calculate the inter-annotator-agreement (IAA) using the pair-wise average $F$-score \cite{Hripcsak05} of instances and tokens. An instance is a single composite annotation unit which consists of one or more tokens, e.g., \textit{``eingeschränkte Nierenfunktion''} (limited renal function) denotes an instance with two (German) tokens, \textit{``eingeschränkte''} and \textit{``Nierenfunktion''}.

The agreement subset consists of 20 text segments with the largest amount of automatic annotations for each of four guidelines (HCC, CLL, Pancreatic cancer, Psycho-oncology) and 20 random-sampled text segments, resulting in
40 
agreement documents with a size of approx. 19k 
tokens annotated by all annotators. We excluded the gene category from the IAA analysis, due to an apparently large number of false positive pre-annotations. 
The IAA achieved an average $F$-score of 0.742 on instances and 0.758 on tokens. Furthermore, we calculated micro-averaged precision and recall values for the automated annotation results, using the complete set of manually reviewed annotations as  gold standard. The results are depicted in Table \ref{tab:evaluation}.

In another annotation study of diagnoses, symptoms and findings on the Jena part of the 3000PA corpus, average $F$-score values converged in the range of around 0.7--0.8 for typical clinical entities as well, e.g., anatomy or disorders in comparison to diagnoses (approx.~0.7), also for pre-annotations. 
The low IAA value of \textit{Physiology} is similar to the IAA of 0.5 on the symptoms category of the named study \cite{Lohr20}. 
The UMLS category \textit{Living Beings} contains a lot of information si\-mi\-lar to personal health information. The average IAA value of around 0.9 is similar to average va\-lues of an annotation study for the anonymization of German discharge summaries ($F$-score $>$ 0.95) \cite{Kolditz19}.


\section{Discussion \& Limitations of this Study}\label{sec:discussion}

While the initial results of the information extraction pipelines we employed are promising, there is much room for improvement. The extraction of genes suffers from a large number of false positives, as there are many common German words (e.g., \textit{``gilt''}, \textit{``dar''}) and three-letter-acronyms (e.g., \textit{``CLL''}, \textit{``HCC''}) with strings identical with gene names in our large dictionary (around~562k entries). Thus, supplying an improved gene tagger which balances German lexical noise with advanced capabilities of
gene taggers for English texts will be a desideratum of future research.

The German UMLS has a number of issues, which severely affect our dictionary-based entity extraction pipelines. 
First and foremost, its vocabulary size is extremely limited. The English UMLS contains over 6.5M entries and the Spanish one around 750k, whereas there are only around 234k entries in the German version (3.6\% coverage of the English version). Recently introduced drugs are missing in the UMLS \textit{Chemistry} category, so a more up-to-date dictionary of drug names is also needed for future work. Moreover, the surface representation of German umlauts is notoriously inconsistent in UMLS, e.g., \textit{``ä''} is sometimes transcribed as \textit{``ae''} or even simplified as \textit{``a''}, as in \textit{``eingeschraenkte Nierenfunktion''}, which results in an increasing false negative rate. 
All of these factors contribute to rather low recall values, as shown in Table \ref{tab:evaluation}.

The accuracy of dictionary matches further decreases due to inconsistent handling of compounds throughout the corpus. For instance,  \textit{``Pankreaskarzinompatienten''} (patients with pancreas carcinoma) would not be detected as an entity, whereas hyphen-connected \textit{``Pankreaskarzinom\underline{-}\underline{P}atienten''} would, yielding two entities (\textit{Disorders} and \textit{Living Beings}), respectively. In this case, we would choose to annotate the whole compound as \textit{Living Beings} to avoid annotation on a subword level, which could be addressed using a more finely adapted tokenization algorithm.
While precision and recall of the rule-based TNM extraction approach are high on \thiscorpus, 
one has to be careful as certain TNM expressions can cause context-dependent semantic ambiguities. For instance, \textit{``V1''} and \textit{``V2''} are valid TNM components referring to venous invasion, but are also detected in the \textsc{WikiWarsDE} corpus referring in this context to German missiles from World War II.
 
\vspace{-3pt}
\section{Conclusion}\label{sec:conclusion}

We presented \thiscorpus, one of the currently largest corpora composed of German medical texts, assembled from the CPGs in oncology and equipped with rich structure information and metadata. We applied 
information extraction pipelines to extract a variety of named entity classes. Despite the limitations we discussed, the information extracted so far can be of immediate use to enable semantic search functionalities in the guideline app \cite{Seufferlein2019}, precision medicine search engines \cite{Faessler20} or in clinical decision support systems \cite{schapranow2015medical}. 

Our results indicate that \thiscorpus\ shares many characteristics with existing clinical text corpora. This can facilitate the development of machine learning-based NLP algorithms for German clinical text. \citet{Beam2020} suggest that combining corpora covering different parts of medical terminology can improve the utility of trained word embeddings.
In addition to the German documents discussed in this work, some of the GGPO guidelines have an additional English version, which could be used to construct parallel corpora for research in multilingual medical NLP.

Extending our work to clinical guidelines from other medical specialities besides oncology will be a straightforward way to extend the volume of the corpus, provided that the document structures can be harmonized across medical societies. However, as most CPGs are distributed as PDF documents, extraction of the plain text content from these can result in quality issues not encountered in this work.

The structured metadata of the corpus provide ample opportunities for future research. For instance, the corpus can be used as a resource for evidence-based medicine summarization, as it contains mappings from literature references to recommendation statements and evidence levels. As we plan to create future versions of the corpus based on updated guideline versions, the extracted concepts can also be used to track changes in CPGs, like the emergence of new treatments and other changes in recommended clinical practice. We envision to combine information extracted from scientific articles, such as study reports, or clinical trial registers with information from CPGs to automatically detect whether these CPGs might be outdated given changes in the underlying evidence base.

In addition to the existing annotations for a wide selection of UMLS semantic types, we can ea\-si\-ly extend the employed pipelines with different dictionaries, e.g., derived from other subsets of the German part of the UMLS, more comprehensive official lists of drug names, or the German version of the \textit{International Classification of Diseases}. 

\vspace{2mm}
We make \thiscorpus\ available for researchers under the conditions of a Data Use Agreement.
For instructions on how to access the corpus and the human annotated data see: \hyperlink{https://www.leitlinienprogramm-onkologie.de/projekte/ggponc-english/}{\url{https://www.leitlinienprogramm-onkologie.de/projekte/ggponc-english/}}. The code to reproduce our experiments is available at: 
\hyperlink{https://doi.org/10.5281/zenodo.4067994}{\url{https://doi.org/10.5281/zenodo.4067994}}.

\begin{small}
\vspace*{+3pt}
\textbf{Acknowledgments.} This work was partially supported by the German Federal Ministry of Research and Education (BMBF) under grants (01ZZ1802H, 01ZZ1803G). We thank our annotators, André Scherag and Danny Ammon from the Jena University Hospital, and all colleagues from the \textsc{HiGHmed} and \textsc{SMITH} consortia for their constant support.
\vspace*{-10pt}
\end{small}

\bibliographystyle{acl_natbib}
\bibliography{refs}

\end{document}